\documentclass[a4paper, 12pt]{article}

\usepackage[T1]{fontenc}

\usepackage{amsmath}
\numberwithin{equation}{section}
\usepackage{graphicx}
\usepackage{url}
\usepackage{amsbsy,amssymb} 
\usepackage{amsmath}
\usepackage{abstract}
\usepackage{hyperref}
\usepackage{color}
\usepackage{authblk}

\usepackage{fancyvrb}

\newcommand{\codeMark}[1]{{\protect\Verb+#1+}} 
\newcommand{\classMark}[1]{{\protect\Verb+#1+}} 

\newcommand{\dirFileMark}[1]{{\it #1}}

\title{\codeMark{TreeGen} - a Monte Carlo generator for data frames.}
\author[1]{A. Niemczynowicz}
\author[2]{G. Bia\l{}osk\'{o}rska}
\author[3]{J. Nie\.{z}urawska-Zaj\k{a}c}
\author[2,4]{R.A. Kycia\thanks{Corresponding author's e-mail: kycia.radoslaw@gmail.com}}
\affil[1]{University of Warmia and Mazury \\ Faculty of Mathematics and Computer Science \\ S\l{}oneczna 54, 10-710 Olsztyn, Poland}
\affil[2]{Cracow University of Technology \\ Faculty of Materials Science and Physics \\ Warszawska 24, 31-155 Krak\'ow, Poland}
\affil[3]{Faculty of Finance and Management, WSB University in Toru\'{n}, M\l{}odzie\.{z}owa 31a, 87-100 Toru\'{n}, Poland}
\affil[4]{Masaryk Univeristy \\ Department of Mathematics and Statistics \\ Kotl\'{a}\v{r}sk\'{a} 267/2, 611 37 Brno, The Czech Republic}

\date{}
\begin{document}
\maketitle
\begin{abstract}
\noindent
The typical problem in Data Science is creating a structure that encodes the occurrence frequency of unique elements in rows and relations between different rows of a data frame. We present the probability tree abstract data structure, an extension of the decision tree, that facilitates more than two choices with assigned probabilities. Such a tree represents statistical relations between different rows of the data frame. The Probability Tree algorithmic structure is supplied with the Generator module that is a Monte Carlo generator that traverses through the tree. These two components are implemented in \codeMark{TreeGen} Python package. The package can be used in increasing data multiplicity, compressing data preserving its statistical information, constructing hierarchical models, exploring data, and in feature extraction.
\end{abstract}


\section{Introduction}
In typical Data Science situations, the categorical data are packed in a Data Frame data structure present in the languages popular in this community, like Python \cite{DataFramePython} or R \cite{DataFrameR}. For an example a set of answers to some query with a typical output presented in Tab. \ref{Tab_Data}. Each column contains an answer to the specific query, and each row contains the record from a specific subject. In the end, the data frame contains categorical data or the data that can be converted to such data, e.g., by data binning.
\begin{table}[htb!]
\begin{center}
\begin{tabular}{|c|c|c|c|}
\hline
\textbf{P1.1} & \textbf{P1.2} & \textbf{P1.3} & \textbf{\ldots }\\ \hline
1 & 2 & 3 & \ldots \\ \hline
5 & 4 & 4 & \ldots  \\ \hline
2 & 2 & 2 & \ldots  \\ \hline
\ldots & \ldots & \ldots & \ldots  \\ \hline
\end{tabular}
\label{Tab_Data}
\caption{Example data. The first row contains the labels of columns (questions).}
\end{center}
\end{table}

Our main motivations to consider this situation are a management and sociological data analysis. In applications there is sometimes a need for generation of additional data with similar statistics, visualize relationship between columns. This could help to extract features and prepare a model of the phenomena described by the data.

The statistical relations between these data can be visualized in the tree-like structure, which is an extension of the decision tree \cite{DecisionTree1, DecisionTree2, DecisionTree3} to multiple choices of alternatives. Such structures are related to Markovian-chain tree (stochastic tree) \cite{Markov, Markov2} and can be used to decision making, e.g., in Medicine \cite{StochasticTree}, Biology \cite{DecisionTree4}, or Management \cite{StochasticTree2}. In applications, there are two opposite spectra in use: on one side the (binary) decision trees, and full Markov graph on the other side. Therefore there is a need for the not necessary binary tree structure that is optimized to store probability of a specific node.

We will call this tree the probability tree. The node of the tree contains:
\begin{itemize}
 \item {The label of the column, e.g., $P1.1$}
 \item {The list of node data that contains:}
 \begin{itemize}
  \item {The value, e.g., $1$;}
  \item {The conditional probability, e.g., $P(P1.2 | P1.1, {value}=\ldots)$;}
  \item {The sub-tree that contains the tree for remaining columns in the Data Frame.}
 \end{itemize}
\end{itemize}

\begin{figure}[htb!]
  \centering
 \includegraphics[width=0.7\textwidth]{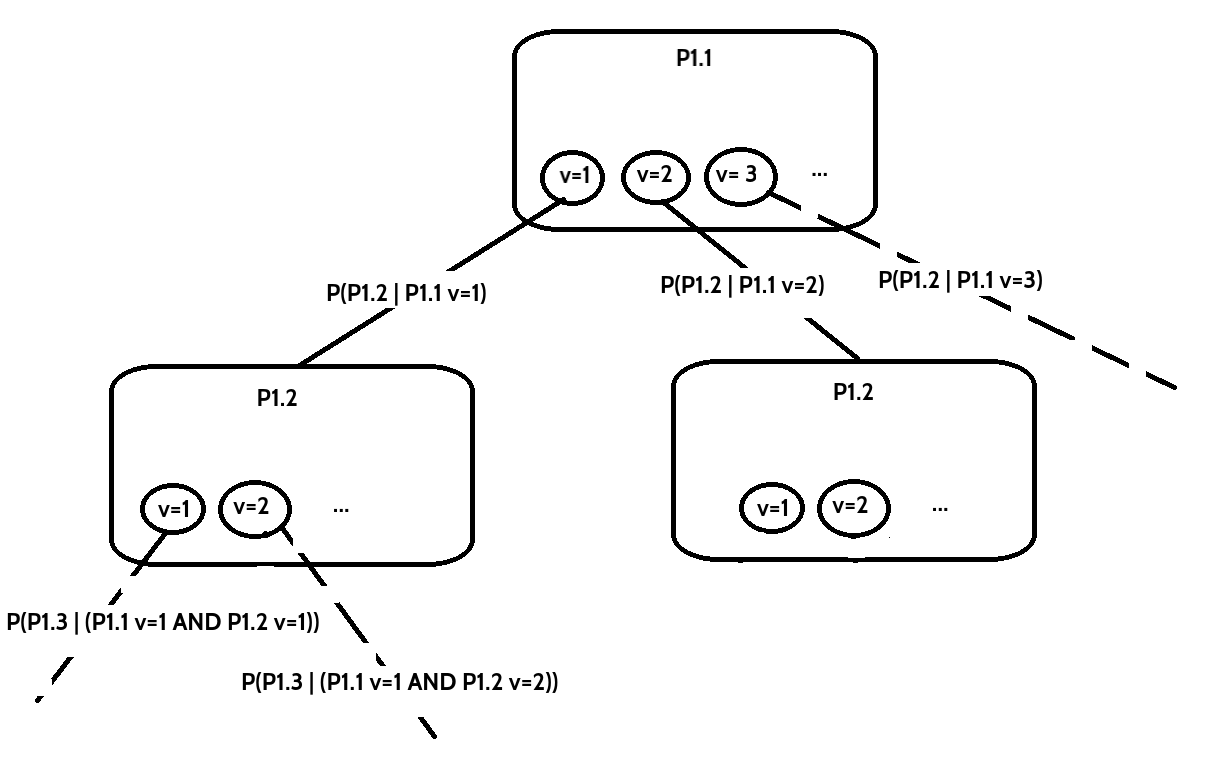}
 \label{Rys.TreeVisualization}
 \caption{A sketch of the probability tree for the data from Tab. \ref{Tab_Data}. Each node contains the column name and its value, and on the vertices the conditional probabilities are marked. For example $P(P1.3 | ([P1.1, v=1] ~ AND ~ [P1.2, v=2]))$ is the conditional probability of the data in column $P1.3$ under the conditions that $P1.1$ has vale $v=1$ and $P1.2$ has value $v=2$.}
\end{figure}

The structure is a variation of the well-known tree structure \cite{Cormen}. A sketch of such a structure is presented in Fig. \ref{Rys.TreeVisualization}.
This structure can be seen as a way to compress the data form the data frame when the statistical properties are the only essential information required. The main idea behind it is the use of the Bayesian inference between different columns in data frame. Therefore such data reduction can be called the 'Bayesian compressing' of categorical data. However, this is not the same as a Bayesian compressing idea for Deep Learning \cite{BayesianCompressing}. Moreover, when the data carry some model then such tree is efficient hierarchical model representation known well from Bayesian analysis \cite{Bayesina1}.

When traversing the probability tree from the root to the leaf along the edges with prescribed probabilities (frequencies of occurrence of values), such structure generates records of values with the same statistics as in the original Data Frame. Such a walk along the tree can be done using a simple Monte Carlo method. Therefore, the Probability Tree can be used as an ADS describing input data for Monte Carlo generators.

The paper contains the description of the probability tree structure implementation and the Generator that yields records using Monte Carlo generation on this tree. The implementation is provided in the pseudocode and specific implementation is provided in Python programming language since it is a popular choice for Data Analysis. The object-oriented paradigm is used.

The paper is organized as follows: In the next section, the description of the \classMark{ProbabilityTree} and \classMark{Generator} classes are given. Then the prerequisites and simple use of the package are presented. Finally, tests of the generator are provided.

\section{Probability Tree class}\label{ProbabilityTree}
In this section, the class \classMark{ProbabilityTree} along with the helper classes is presented. We start from the \classMark{Node} class.

\subsection{Nodes}
\begin{figure}[htb!]
  \centering
 \includegraphics[width=0.5\textwidth]{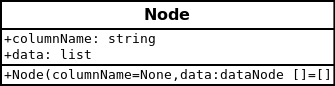}
 \caption{The \classMark{Node} class.}
 \label{Rys.NodeClass}
\end{figure}

The basic building block of a Probability Tree is the \classMark{Node} class that is presented in Fig. \ref{Rys.NodeClass}.
The name of the fields are as follows:
\begin{itemize}
 \item {\codeMark{columnName} - the name of the column of the \classMark{DataFrame} for this node.}
 \item {\codeMark{data} - the list that contains \classMark{dataNode} elements.}
\end{itemize}

\begin{figure}[htb!]
  \centering
 \includegraphics[width=0.5\textwidth]{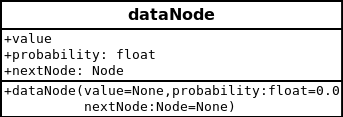}
 \caption{The \classMark{dataNode} class.}
 \label{Rys.dataNodeClass}
\end{figure}

The \codeMark{data} field contains the \classMark{dataNode} objects stored in the list. This node is presented in Fig. \ref{Rys.dataNodeClass}.
The fields have the following meaning:
\begin{itemize}
 \item {\codeMark{value} - given value for the column of the \classMark{DataFrame}.}
 \item {\codeMark{probability} - the probability (frequency of occurrence) of the unique value in the column.}
 \item {\codeMark{nextNode} - sub-tree created from the \classMark{DataFrame} with fixed value.}
\end{itemize}

These nodes are building blocks of a tree in the core class \classMark{ProbabilityTree} described in the next subsection.

\subsection{ProbabilityTree}
\begin{figure}[htb!]
  \centering
 \includegraphics[width=0.5\textwidth]{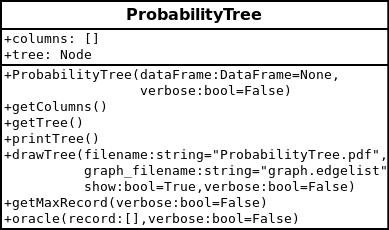}
 \caption{The \classMark{ProbabilityTree} class.}
 \label{Rys.ProbabilityTreeClass}
\end{figure}

The \classMark{ProbabilityTree} class is presented in Fig. \ref{Rys.ProbabilityTreeClass}.
The fields are
\begin{itemize}
 \item {\codeMark{columns} - is the list that contains the columns of the \classMark{DataFrame} class with proper ordering that matches the labelling of tree levels. The method \codeMark{getColumns} returns this field.}
 \item {\codeMark{tree} - contains the whole Probability Tree. The method \codeMark{getTree} returns this field.}
\end{itemize}

The constructor \codeMark{ProbabilityTree(dataFrame, verbose)} takes the \classMark{DataFrame} object in \codeMark{dataFrame} variable and construct the Probability Tree. The variable \codeMark{verbose}, which occurs also in the other methods prints the diagnostic data. The method uses helper method that recursively construct the tree from the \classMark{DataFrame}. The pseudocode is as follows:
\begin{itemize}
 \item {Create \classMark{Node} object and assign \codeMark{columnName} to the first column.}
 \item {Calculate the probabilities (frequencies) of the unique values in the first column in the \codeMark{dataFrame} variable.}
 \item {Order unique elements with respect to the decreasing probability. This will speed up searching the maximal elements in the node in the other methods.}
 \item {For a given fixed value of the first column:}
 \begin{itemize}
  \item {Calculate recursively subtree for other columns that the first column has fixed value.}
  \item {Assign \codeMark{value}, \codeMark{probability} and \codeMark{nextNode} to a new \classMark{dataNode} object.}
  \item {Append \classMark{dataNode} object to the \codeMark{data} list.}
 \end{itemize}
\end{itemize}

Similar recursive walk is used in the other methods described below.

The method \codeMark{printTree} print tree using the pre-order recursive walk. 

\begin{figure}[htb!]
  \centering
 \includegraphics[width=0.9\textwidth]{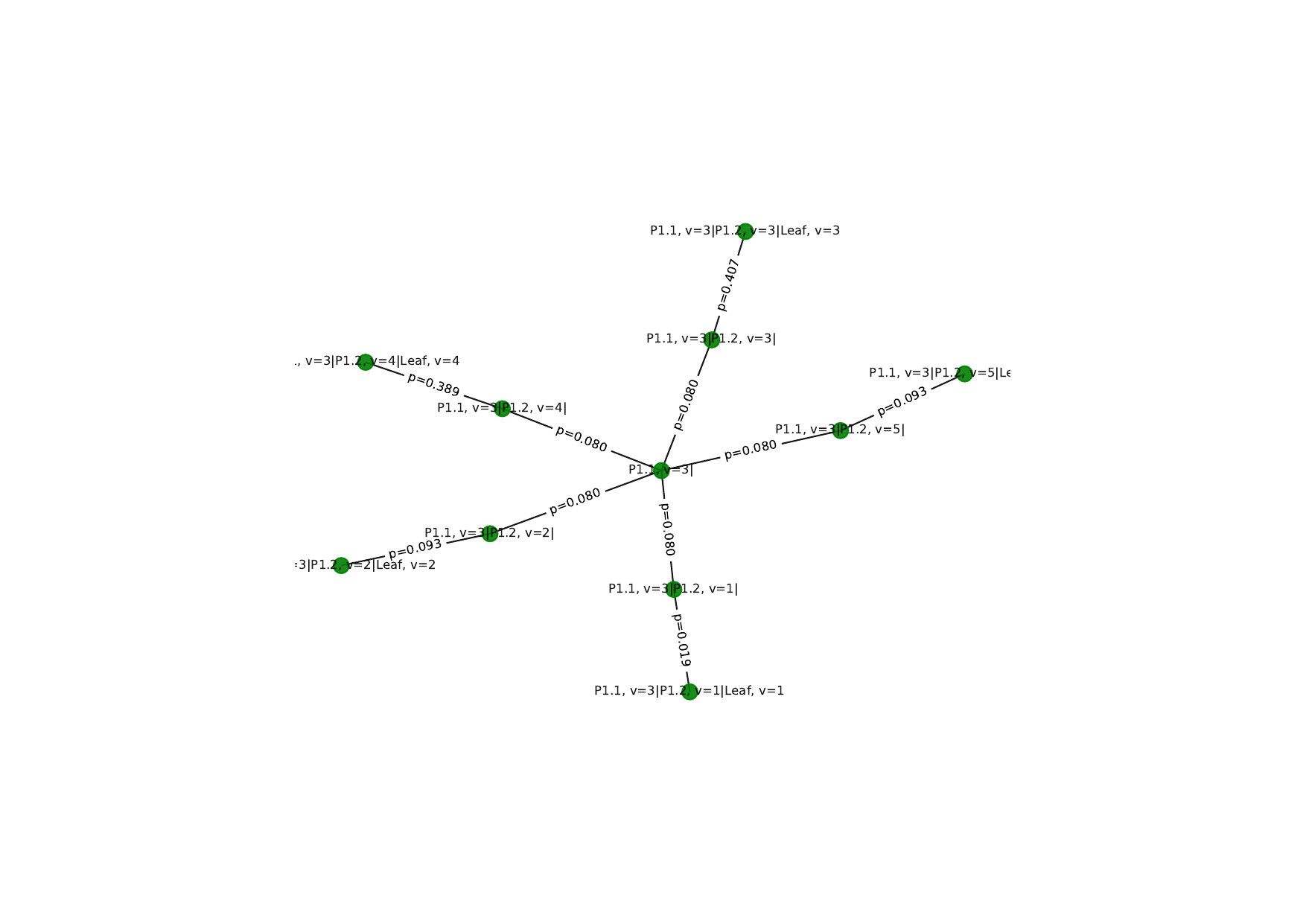}
 \caption{One of the component of a \classMark{ProbabilityTree} obtained from a \classMark{DataFrame} by the method \codeMark{drawTree}.}
 \label{Rys.ProbabilityTreeGraph}
\end{figure}

Similarly, the method \codeMark{drawTree(filename, graph\_filename, show, verbose)} saves the graphs visualizing tree in the file of variable \codeMark{fielname} and displays it when \codeMark{show} is set to \codeMark{True}. In the Python implementation the NetworkX library \cite{networkx} is used and the graph data is saved in the file \codeMark{graph\_filename}. The method constructs the list of edges by a recursive walk through the tree and picks up the probability of the edge. The vertices are named by the path (column name, value = ... ) by which it is reached from, e.g., $P1.1, v=1 | P1.2 v=2 | \ldots$, which ends with $Leaf$ sentence if it is the leaf of the tree - the last node. Since the root node can be interpreted as an insertion point to the graphs that holding connected components for different values in the first column of a data, so the extraction of such subtrees can be performed by selecting connected components of the full graph. The graph, containing the whole tree, is divided into connected components associated with the unique values of the first column of the \classMark{DataFrame} object used to construct the \classMark{ProbabilityTree}. An example of such connected component is presented in Fig. \ref{Rys.ProbabilityTreeGraph}. It represents the splitting the original data into columns using Bayesian inference.

The next method \codeMark{getMaxRecord} recursively traverse the Probability Tree and returns the tuple containing the lists of values with maximal probabilities at vertices (in the order returned by \codeMark{getColumns}) and probabilities of these values.

The final public method is \codeMark{oracle(record, verbose)}, that takes in \codeMark{record} the list of values (assumed to be ordered in the way returned by \codeMark{getColumns}) and checks if the \classMark{ProbabilityTree} contains it. The list can be shorter than the hight of the tree, and then the method checks if the initial sequence is in the tree.

\begin{figure}[htb!]
  \centering
 \includegraphics[width=0.4\textwidth]{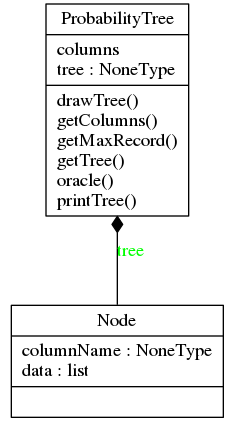}
 \caption{Dependence between \classMark{ProbabilityTree} and \classMark{Node} classes.}
 \label{Rys.ProbabilityTreeCompositionDiagram}
\end{figure}

The composition diagram is presented in Fig. \ref{Rys.ProbabilityTreeCompositionDiagram}.

\section{Generator class}\label{Generator}
\begin{figure}[htb!]
  \centering
 \includegraphics[width=0.5\textwidth]{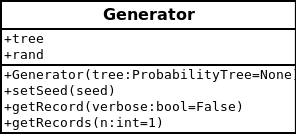}
 \caption{The class \classMark{Generator}.}
 \label{Rys.GeneratorClass}
\end{figure}

The \classMark{Generator} class is presented in Fig. \ref{Rys.GeneratorClass}.
The class contains two public attributes
\begin{itemize}
 \item {\codeMark{tree} - contains the \classMark{ProbabilityTree} object that is used for generation.}
 \item {\codeMark{rand} - contains the random number generator used for Monte Carlo generation.}
\end{itemize}

The constructor takes the \classMark{ProbabilityTree} object and initializes the internal random number generator. The seed can be reset by the \codeMark{setSeed} method.

Two core methods are \codeMark{getRecord} and \codeMark{getRecords}. The first one recursively walks through the tree according to the conditional probabilities of the nodes, and returns the data record as \classMark{DataFrame}. The next node is selected when the generated (pseudo)random number is compared with the cumulative distribution of all probabilities in the given node, see Fig. \ref{Rys.TreeVisualization}.

The second method returns \codeMark{n} records. The values in the node of the tree are selected using a simple Monte Carlo algorithm:
\begin{itemize}
 \item {Generate a random number from the uniform distribution on the unit interval.}
 \item {Iterate over \codeMark{Node:data} until cumulative distribution for a given value is not less than the random number. Then select the next node according to the obtained probability and pick the corresponding value associated with this probability.}
\end{itemize}

\section{Requirements and Usage}\label{Usage}
The generator is contained in the Python package \dirFileMark{TreeGen} which is freely available at \cite{implementation}. The installation can be done by copying the \dirFileMark{TreeGen} directory into the standard Python 3 library directory, which is present in the $PYTHONPATH$ environmental variable. The package requires the following elements:
\begin{itemize}
 \item {Python 3 \cite{Python} as a runtime environment.}
 \item {Pandas library \cite{Pandas} for \classMark{DataFrame} class.}
 \item {Matplotlib library \cite{Matplotlib} for graphics.}
 \item {NetworkX library \cite{networkx} for drawing \classMark{ProbabilityTree} structure.}
 \item {Doxygen software \cite{Doxygen} for automated generation of documentation.}
\end{itemize}

Its import can be done as
\codeMark{from TreeGen.Generator import *}.\\
It imports both \classMark{ProbabilityTree} as well as \classMark{Generator} modules with the classes of the same names.

Having defined a \classMark{DataFrame} object as, e.g., \codeMark{dataFrame}, the \codeMark{ProbabilityTree} object can be created by 
\codeMark{tree = ProbabilityTree(pd)}. Finally, the generator object can be created by \codeMark{gen = Generator(tree)}.

An example usage can be as follows:
\begin{verbatim}
import pandas as pd
import TreeGen.Generator as G
#read data from Data.xls file:
df = pd.read_excel('Data.xls')
#create ProbabilityTree:
tree = G.ProbabilityTree( df )
#show the data stored in ProbabilityTree:
tree.drawTree(verbose=True, show = False)
#create MC Generator based on ProbabilityTree:
gen = G.Generator(tree)
#Generate 1000 records:
genData = gen.getRecords(1000)        
print( genData.head() )
\end{verbatim}

A more advanced scenario of using these classes is provided by \dirFileMark{Main.py} file supplied with the package.

In the next section, some validation tests are presented.
\section{Validation}\label{Validation}
In the test, the two-column dataset of $672$ records was used. We generate $1000$ records and then compare statistics of the first columns of the data and generated records. 

\begin{figure}[htb!]
 \centering
 \includegraphics[width=0.4\textwidth]{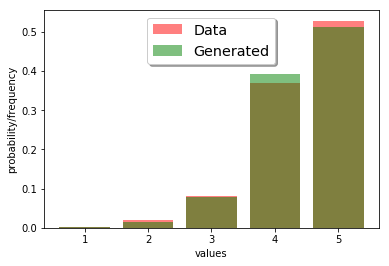}
 \includegraphics[width=0.4\textwidth]{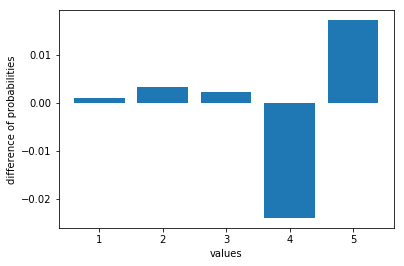}
 \caption{Comparison of frequency of occurrence of unique values (left) for the data ($672$ records) and generated data ($1000$ records) for the first column. The difference between these two frequencies are visualized in the right figure.}
 \label{Rys.SingleRunComparisonP11}
\end{figure}

\begin{figure}[htb!]
 \centering
 \includegraphics[width=0.4\textwidth]{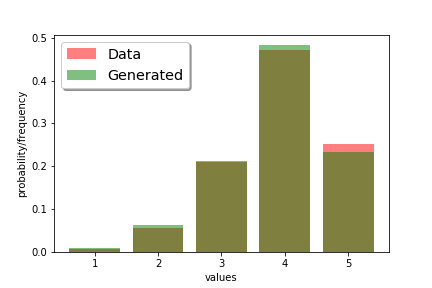}
 \includegraphics[width=0.4\textwidth]{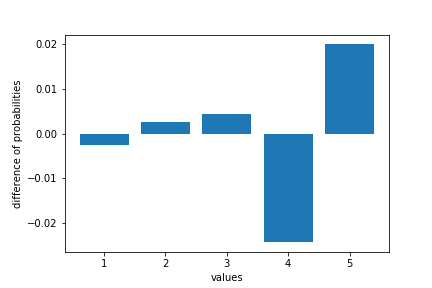}
 \caption{Comparison of frequency of occurrence of unique values (left) for the data ($672$ records) and generated data ($1000$ records) for the second column. The difference between these two frequencies are visualized in the right figure.}
 \label{Rys.SingleRunComparisonP12}
\end{figure}

The unique values in the first column are $1,2,3,4,5$. The comparison of frequencies of the first column of the data ($672$ records) and the results obtained from the generator ($1000$ records) was compared.  An example run is presented in Fig. \ref{Rys.SingleRunComparisonP11} for the first column of the data frame, and in Fig. \ref{Rys.SingleRunComparisonP12} for the second column of the data. It shows a good agreement of the generated records with data.

\begin{figure}[htb!]
 \centering
 \includegraphics[width=0.6\textwidth]{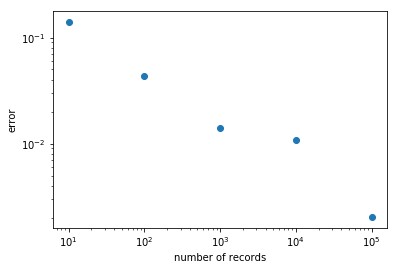}
 \caption{Convergence of the generated records to the data. The error is the $L_{1}$ norm of difference of frequencies for unique values of first column of the data and for generated data.}
 \label{Rys.Convergence}
\end{figure}

In the second test we showed the usual Monte Carlo-type convergence of frequencies for generated items to the data in the first column when we increase statistics of generated records. Fig. \ref{Rys.Convergence} shows a decrease in the order of error per increase of statistics by two orders of magnitude per $100$-times increase in statistics. It is typical behavior for Monte Carlo methods.

\section{Impact and Conclusions}\label{Conclusions}
The Probability Tree abstract data structure, along with its implementation, was presented. It can store the frequency relationships of the data stored in the Data Frame. This structure can visualize relationships between the data and generate data with the same probability characteristics using the supplied Monte Carlo generator. This information can be used to construct a relationship model in the data, and hence it is a useful tool for data exploration. The test of a Python implementation was performed, and the correct Monte Carlo-type convergence was presented on example data.

The proposed library can be used in Data Science and Machine Learning as well as in data exploration and feature discovery.

\section*{Acknowledgments}
The work of AN, JN-Z and RK was supported in part by NAWA founded grant 'The International Academic Partnership for Generation Z'. RK was also supported by GACR grant GA19-06357S and the MUNI/A/0885/2019 grant of Masaryk University.




\end{document}